\documentclass{article}

\usepackage{microtype}
\usepackage{graphicx}
\usepackage{subfigure}
\usepackage{booktabs} 

\usepackage{hyperref}

\usepackage[accepted]{icml2019}

\icmltitlerunning{Learning Optimal Fair Policies}

\newtheorem{thm}{Theorem}
\usepackage{natbib}
\usepackage{tikz}
\usepackage{epsfig,amssymb,amsmath,ifthen,comment}

\newcommand{\E}{\mathbb{E}}
\DeclareMathOperator{\pa}{pa}

\DeclareMathOperator*{\argmax}{arg\,max}

\begin{document}

\twocolumn[
\icmltitle{Learning Optimal Fair Policies}



\icmlsetsymbol{equal}{*}

\begin{icmlauthorlist}
\icmlauthor{Razieh Nabi}{to}
\icmlauthor{Daniel Malinsky}{to}
\icmlauthor{Ilya Shpitser}{to}
\end{icmlauthorlist}

\icmlaffiliation{to}{Department of Computer Science, Johns Hopkins University, Baltimore, MD, USA}

\icmlcorrespondingauthor{Razieh Nabi}{rnabi@jhu.edu}

\icmlkeywords{Fair Policies, Policy Learning, Causal Inference, Mediation Analysis, Algorithmic Fairness, Dynamic Treatment Regimes, Optimal Policies}

\vskip 0.3in
]



\printAffiliationsAndNotice{}  

\begin{abstract}
Systematic discriminatory biases present in our society influence the way data is collected and stored, the way variables are defined, and the way scientific findings are put into practice as policy. Automated decision procedures and learning algorithms applied to such data may serve to perpetuate existing injustice or unfairness in our society. In this paper, we consider how to make optimal but fair decisions, which ``break the cycle of injustice'' by correcting for the unfair dependence of both decisions and outcomes on sensitive features (e.g., variables that correspond to gender, race, disability, or other protected attributes). 
We use methods from causal 
inference and constrained optimization to learn optimal policies in a way that addresses 
multiple potential biases which afflict data analysis in sensitive contexts,
extending the approach of \citet{NabiShpitser18Fair}.
Our proposal comes equipped with the theoretical guarantee that the chosen fair policy will 
induce a joint distribution for new instances
that satisfies given fairness constraints.
We illustrate our approach with both synthetic data and real criminal justice data.


\end{abstract}

\section{Introduction} 
Making optimal and adaptive intervention decisions in the face of uncertainty is a central task in precision medicine, computational social science, and artificial intelligence. In healthcare, the problem of learning optimal policies 
is studied under the heading of \emph{dynamic treatment regimes} \cite{Moodie13DTR}.  The same problem is called \emph{reinforcement learning} in artificial intelligence \cite{sutton98reinforcement}, and \emph{optimal stochastic control} \cite{bertsekas96neuro} in engineering and signal processing. In all of these cases, a policy (a function of historical data to some space of possible actions, or a sequence of such functions) is chosen to maximize some pre-specified outcome quantity, which might be abstractly considered a \emph{utility} (or \emph{reward} in reinforcement learning). Increasingly, ideas from optimal policy learning 
are being applied in new contexts. In some areas, particularly socially-impactful settings like criminal justice, social welfare policy, hiring, and personal finance, it is essential that automated decisions respect principles of fairness since the relevant data sets include potentially sensitive attributes (e.g., race, gender, age, disability status) and/or features highly correlated with such attributes, so ignoring fairness considerations may have socially unacceptable consequences.  A particular worry in the context of automated sequential decision making is ``perpetuating injustice,'' i.e., when maximizing utility maintains, reinforces, or even introduces unfair dependence between sensitive features, decisions, and outcomes. Though there has been growing interest in the issues of fairness in machine learning \cite{pedreshi08discrimination,feldman15certifying,hardt16equality,kamiran13quantifying,corbett2017algorithmic,jabbari16fair,kusner17counterfactual,zhang18fairness,mitchell18speak,Zhang17causal}, so far methods for 
optimal policy learning subject to fairness constraints
have not been well-explored. 

As a motivating example, we consider a simplified model for a children's welfare screening program, recently discussed in \cite{chouldechova2018case, Hurley18child}. A hotline for child abuse and neglect receives many thousands of calls a year, and call screeners must decide on the basis of calculated risk estimates what action to take in response to any given call, e.g., whether or not to follow up with an in-person visit from a caseworker. The idea is that only cases with substantial potential risk to the child's welfare should be prioritized. The information used to determine the calculated risk level and thereby the agency's action includes potentially sensitive features, such as race and gender, as well as a myriad of other factors such as perhaps whether family members receive public assistance, have an incarceration history, record of drug use, and so on. Though many of these factors may be predictive of subsequent negative outcomes for the children, there is a legitimate worry that both risk calculations and policy choices based on them may depend on sensitive features in inappropriate ways, and thereby lead
to unfair racial disparities in the distribution of families investigated, and perhaps separated, by child protective services.

Learning high-quality policies that satisfy fairness constraints is difficult due to the fact that multiple sources of bias may occur in the problem simultaneously. One kind of bias, which we call \emph{retrospective bias}, has its origin in the historical data used as input to the policy learning procedure.  This data may reflect various systematic disparities and discriminatory historical practices in our society, including prior decisions themselves based on poor data.  Algorithms trained on such data can 
maintain these inequities.  Furthermore, decision making algorithms may suffer from what we call \emph{prospective} sources of bias. For instance, suppose the functional form of the chosen decision rule explicitly depends on sensitive features in inappropriate ways.  In that case, making decisions based on the new decision rule may perpetuate existing disparities or even introduce disparities that were previously absent.  Avoiding this sort of bias may involve imposing non-trivial restrictions on the policy learning procedure. Finally, learning high-quality policies from observational data requires dealing with \emph{confounding bias}, where associations between decision and reward cannot be used directly to assess decision quality due to the presence of confounding variables, as well as \emph{statistical bias} due to the reliance on misspecified statistical models.  Policy learning algorithms that respect fairness constraints must address all of these sources of bias.

In this paper, we use tools from mediation analysis and causal inference to formalize fairness criteria as constraints on certain impermissible causal pathways from sensitive features to actions or outcomes \cite{NabiShpitser18Fair}. Moreover, we describe how all the aforementioned biases can be addressed by a novel combination of methods from causal inference, constrained optimization, and semiparametric statistics. Our main theoretical result illustrates in what sense enacting fair policies can ``break the cycle of injustice'':
we show how to learn policies such that the joint distribution induced by these policies (in conjunction with reward/utility mechanisms outside the policy-maker's control) will satisfy specified fairness constraints while remaining ``close'' to the generating distribution.
To our knowledge, this paper constitutes the first attempt to integrate algorithmic fairness and policy learning with the possible exception of \citet{jabbari16fair}, which addressed what we call prospective bias in the context of Markov Decision Processes. 


To precisely describe our approach, we must introduce some necessary concepts and tools from causal inference and policy learning. Then, we summarize the perspective on algorithmic fairness in prediction problems from \citet{NabiShpitser18Fair}, and adapt this framework to learning optimal fair policies. We illustrate our proposal via experiments on synthetic and real data.

\section{Notation and preliminaries} 
\label{sec:prelim}

Consider a multi-stage decision problem with $K$ pre-specified decision points, indexed by $k = 1, \ldots, K$.  Let $Y$ denote the final outcome of interest and $A_k$ denote the action made (treatment administered) at decision point $k$ with the finite state space of ${\cal A}_k$. Let $X$ denote the available information prior to the first decision, and $Y_k$ denote the information collected between decisions $k$ and $k+1$, ($Y \equiv Y_{K}$). $\overline{A}_k$ represents all treatments administered from time $1$ to $k$; likewise for $\overline{Y}_k$. We combine the treatment and covariate history up to treatment decision $A_k$ into a history vector $H_k$. The state space of $H_k$ is denoted by ${{\cal H}_k}$. Note that while our proposal in this paper applies to arbitrary state spaces,
we
present examples with continuous outcomes and binary decisions for simplicity.

The goal of policy learning is to find
policies that map vectors in ${\cal H}_k$ to values in ${\cal A}_k$ (for all $k$)
that
maximize the expected value of outcome $Y$.
In offline settings, where exploration by direct experimentation is impossible, finding such policies requires reasoning counterfactually, as is common in causal inference.
The value of $Y$ under an assignment of value $a$ to variable $A$
is called a \emph{potential outcome} variable, denoted $Y(a)$. In causal inference, quantities of interest are defined as functions of potential outcomes (also called counterfactuals). Estimating these functions  from observational data is a challenging task, and requires assumptions linking potential outcomes to the data actually observed. Our assumptions can be formally represented using \textit{causal graphs}.  In a directed acyclic graph (DAG), nodes correspond to random variables, and directed edges represent direct causal relationships. As an example, consider the single treatment causal graph of Fig. \ref{fig:triangle}(a). $X$ is a direct cause of $A$, and $A$ is both a direct cause of $Y$ as well as an indirect cause of $Y$ through $M$. A variable like $M$ which lies on a causal pathway from $A$ to $Y$ is called a \textit{mediator}.
For more details on causal graphical models see, e.g., \citet{spirtes01causation} and \citet{pearl09causality}. In what follows let $Z$ denote the full vector of observed variables in the causal model, e.g., $Z = (Y,M,A,X)$ in our Fig. \ref{fig:triangle}(a).

A causal parameter is said to be \emph{identified} in a causal model if it is a function of the observed data distribution $p(Z)$. In causal DAGs, distributions of potential outcomes are identified by the \emph{g-formula}. For background on general identification theory, see \citet{shpitser18identification}.
As an example, the distribution of $Y(a)$ in the DAG in Fig.~\ref{fig:triangle}(a) 
is identified by $\sum_{X,M} p(Y | a,M,X) p(M | a,X) p(X)$.
Note that some causal parameters may be identified even in causal models with hidden (``latent'') variables, typically represented by acyclic directed mixed graphs (ADMGs) \citep{shpitser18identification}. Though we do not apply our methods to hidden variable models here, the general approach and many of the specific learning strategies we propose are applicable in contexts with hidden variables, so long as the relevant parameters are identified.

In our sequential setting,
$Y(\overline{a}_K)$ represents the response $Y$ had the fixed treatment assignment strategy $\overline{A}_K = \overline{a}_K$ been followed, possibly contrary to fact.  The contrast 
$\E[Y(\overline{a}_K)] - \E[Y(\overline{a}'_K)]$, where $\overline{a}_K$ is the treatment history of interest and $\overline{a}'_K$ is the reference treatment history, quantifies the \emph{average causal effect} of $\overline{a}_K$ on the outcome $Y$.

\subsection*{Mediation and path-specific analysis}


One way to understand the mechanisms by which treatments influence outcomes is via mediation analysis.
The simplest type of mediation analysis
decomposes the causal effect of $A$ on $Y$ into a \textit{direct effect} and an \textit{indirect effect} mediated by a third variable. Consider the graph in Fig \ref{fig:triangle}(a): the direct effect corresponds to the path $A \rightarrow Y$, and indirect effect corresponds to the path through $M$: $A \rightarrow M \rightarrow Y$. In the potential outcome notation, the direct and indirect effects can be defined using nested counterfactuals such as $Y(a, M(a'))$ for $a, a' \in \cal A$, which denotes the value of $Y$ when $A$ is set to $a$ while $M$ is set to whatever value it would have attained had $A$ been set to $a'$. Under certain identification assumptions discussed by \citet{pearl01direct}, the distribution of $Y(a, M(a'))$ (and thereby direct and indirect effects) can be nonparametrically identified from observed data by the following formula:
$p(Y(a, M(a')) = \sum_{X, M} p(Y \mid a, X, M) p(M \mid a', X) p(X)$.
More generally, when there are multiple pathways from $A$ to $Y$ one may define various \textit{path-specific effects} (PSEs), which under some assumptions may be nonparametrically identified by means of the \textit{edge g-formula} provided in \citet{shpitser15hierarchy}.  We define a number of PSEs relevant for our examples below.  For a general definition, see \citet{shpitser13cogsci}.

\subsection*{Policy counterfactuals and policy learning}
\label{sec:policy}

Let $f_A = \{f_{A_1}, \ldots, f_{A_K} \}$ be a sequence of decision rules. 
At the $k$th decision point, the $k$th rule $f_{A_k}$ maps the available information prior to the $k$th treatment decision $H_k$ to treatment decision $a_k$, i.e.\ $f_{A_k}: {{\cal H}_k} \mapsto {{\cal A}_k}$.
Given $f_A$ we define the counterfactual response of $Y$ had $A$ been assigned according to $f_A$,
or $Y(f_A)$, by the following recursive definition \citep[cf.][]{robins04SNMM,thomas13swig}:
{\small
\begin{align*}
Y\big(
\big\{ f_{A_k}\big(H_k(f_{A})\big): A_k \in \pa_{\cal G}(Y)\cap A \big\},
\big\{ \pa_{\cal G}(Y) \setminus A \big\}(f_A)
\big).
\end{align*} 
}%
In words: the potential outcome $Y$ had any parent of $Y$ that is in $A$ been set to $f_A$ in response to counterfactual history $H_k$ up to $k$, where this history behaves as if $A$ were set to $f_A$ \textit{and} any parent of $Y$ that is not in $A$, behaves as if $A$ were set to $f_A$. 

\begin{figure*}
\begin{center}
\begin{tikzpicture}[>=stealth, node distance=1.0cm]
    \tikzstyle{format} = [draw, very thick, circle, minimum size=5.0mm,
	inner sep=0pt]
	\begin{scope}
		\path[->, very thick]
			node[format] (c) {$X$}
			node[format, below right of=c] (a) {$A$}		
			node[format, above right of=a] (m) {$M$}
			node[format, below right of=m] (y) {$Y$}
			(c) edge[blue] (a)
			(a) edge[blue] (y)
			(a) edge[blue] (m)
			(m) edge[blue] (y)
			(c) edge[blue] (m)
			(c) edge[blue] (y)
			node[below of=a, yshift=0.1cm, xshift=0.3cm] (l) {$(a)$}	
			;
	\end{scope}
	
	\begin{scope}[xshift=3.0cm] 
		\path[->, very thick]
			node[format] (x) {$X$}
			node[format, below right of=x] (s) {$S$}
			node[format, above right of=s] (m) {$M$}
			node[format, below right of=m] (a) {$A$}
			node[format, above right of=a] (y) {$Y_1$}
			node[format, right of=y] (y2) {$Y_2$}
			(s) edge[blue] (a)
			(s) edge[blue] (m) 
			(m) edge[blue] (a)
			(x) edge[blue] (m)
			(x) edge[blue] (a)
			(a) edge[blue] (y)
			(m) edge[blue] (y)
			(s) edge[blue] (y) 
			(x) edge[blue, bend left=25] (y)
			(x) edge[blue, bend left] (y2)
			(a) edge[blue] (y2)
			(y) edge[blue] (y2)
			(m) edge[blue, bend left] (y2)
			 
			node[below of=a, yshift=0.1cm, xshift=0.0cm] (l) {$(b)$}
		;
	\end{scope}

	\begin{scope}[xshift=8.0cm] 
	\path[->, very thick]
	
	node[format] (x) {$X$}
	node[format, below right of=x] (s) {$S$}
	node[format, above right of=x] (m) {$M$}
	node[format, right of=s] (a1) {$A_1$}
	node[format, right of=m] (y1) {$Y_1$}
	node[  right of=a1] (dots1) {$\ldots$}
	node[  right of=y1] (dots2) {$\ldots$}
	node[format,  right of=dots1] (aK) {$A_K$}
	node[format,  right of=dots2] (yK) {$Y_K$}
	
	(y1) edge[blue] (dots2)
	(a1) edge[blue] (dots2)

	(y1) edge[blue] (dots1)
	(a1) edge[blue] (dots1)

	(dots1) edge[blue, bend left=0] (aK)
	(dots2) edge[blue, bend left=0] (aK)
	(m) edge[blue, bend left=0] (aK)
	(y1) edge[blue] (aK)
	(a1) edge[blue, bend right=15] (aK)
	(s) edge[blue, , bend right=25] (aK)
	(x) edge[blue, bend left=0] (aK)

	(dots1) edge[blue] (yK)
	(dots2) edge[blue] (yK)
	(m) edge[blue, bend left=25] (yK)
	(x) edge[blue, bend left=0] (yK)
	(y1) edge[blue, bend left=15] (yK)
	(s) edge[blue, bend left=0] (yK)
	(a1) edge[blue] (yK)
	(aK) edge[blue] (yK)


	(x) edge[blue] (s)

	(s) edge[blue] (a1)
	(m) edge[blue] (a1)
	(x) edge[blue] (a1)

	(s) edge[blue] (m)
	(x) edge[blue] (m)

	(a1) edge[blue] (y1)
	(m) edge[blue] (y1)
	(x) edge[blue, bend left=0] (y1)
	(s) edge[blue] (y1)

	node[below of=a1, yshift=0.1cm, xshift=0.5cm] (l) {$(c)$}
	;
	\end{scope}

\end{tikzpicture}
\end{center}
\caption{(a) A simple causal DAG, with a single treatment $A$, a single outcome $Y$, a vector $X$ of baseline variables, and a single mediator $M$.
(b) A causal DAG corresponding to our (simplified) child welfare example with baseline factors $X$, sensitive feature $S$, action $A$, vector of mediators (including e.g.\ socioeconomic variables, histories of drug treatment) $M$, an indicator $Y_1$ of whether a child is separated from their parents, and an indicator of child hospitalization $Y_2$. (d) A multistage decision problem, which corresponds to a complete DAG over vertices $X,S,M,A_1,Y_1,\cdots,A_K,Y_K$.
}
\label{fig:triangle}
\end{figure*}
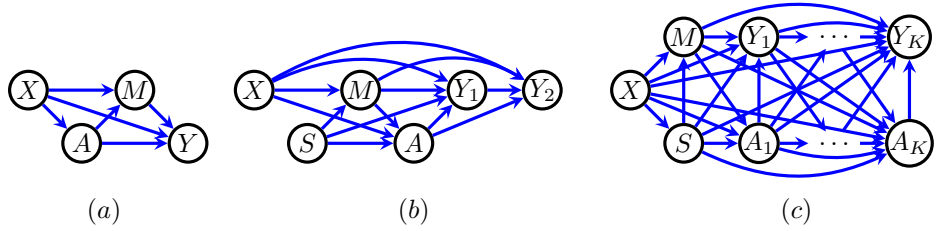

Under a causal model associated with the DAG $\cal G$, the distribution $p(Y(f_A))$, is identified by the following generalization of the g-formula:
{\scriptsize
\begin{align*}
\sum_{Z \setminus \{Y, A \}}   \prod_{V \in Z \setminus A}
p\big(V | \{ f_{A_k}(H_k): A_k \in \pa_{\cal G}(V) \cap A \}, \pa_{\cal G}(V) \setminus A \big).
\end{align*}
}%
As an example, $Y(a = f_A(X))$ in Fig.~\ref{fig:triangle}(a) is defined as $Y(a=f_A(X), M(a=f_A(X),X), X)$, 
and its distribution is identified as
$\sum_{x,m} p(Y | a=f_A(x),M=m,X=x) p(M | a=f_A(x),X=x) p(X=x)$.

Given an identified response to a fixed set of policies $f_A$, we consider search for the optimal policy set $f^*_A$, defined to be one that maximizes $\E[Y(f_A)]$.
Since $Y(f_A)$ is a counterfactual quantity, validating the found set of policies is difficult given only retrospective data, with statistical bias due to model misspecification being a particular worry.  This stands in contrast with online policy learning problems in reinforcement learning, where new data under any policy may be generated and validation is therefore automatic. Partly in response to this issue, a set of orthogonal methods for policy learning have been developed that model different parts of the observed data likelihood function.
Q-learning, value search, and g-estimation are common methods used in dynamic treatment regimes literature for learning optimal policies \cite{Moodie13DTR}. We defer detailed descriptions to later in the paper and the supplement.

\section{From fair prediction to fair policies}
\label{sec:fairpolicy}


\citet{NabiShpitser18Fair} argue that fair inference for prediction requires imposing hard constraints on the prediction problem, in the form of restricting certain path-specific effects. We adapt this approach to optimal sequential decision-making.
A feature of this approach is that the relevant restrictions are user-specified and context-specific; thus we will generally require input from policymakers, legal experts, bioethicists, or the general public in applied settings. Which pathways may be considered impermissible depends on the domain and the semantics of the variables involved. We do not defend this perspective on fairness here for lack of space; please see \citet{NabiShpitser18Fair}
for more details.


We summarize the proposal from \citet{NabiShpitser18Fair} with a brief example, inspired by the aforementioned child welfare case. Consider a simple causal model for this scenario, shown in Fig.\ 1(b). Hotline operators receive thousands of calls per year, and must decide on an action $A$ for each call, e.g., whether or not to send a caseworker. These decisions are made on the basis of a (high-dimensional) vectors of covariates $X$ and $M$, as well as possibly sensitive features $S$, such as race. $M$ consists of mediators of the effect of $S$ on $A$. $Y_1$ corresponds to an indicator for whether the child is separated from their family by child protective services, and $Y_2$ corresponds to child hospitalization (presumably attributed to domestic abuse or neglect). The observed joint distribution generated by this causal model would be $p(Y_1,Y_2,A,M,S,X)$. The proposal from \citet{NabiShpitser18Fair} is that fairness corresponds to the impermissibility of certain path-specific effects, and so fair inference requires decisions to be made from a counterfactual distribution $p^*(Y_1,Y_2,A,M,S,X)$ which is ``nearby'' to $p$ (in the sense of minimal Kullback-Leibler divergence) but where these PSEs are constrained to be zero. They call $p^*$ the distribution generated by a ``fair world.''

Multiple fairness concerns have been raised by experts and advocates in discussions of the child protection decision-making process \cite{chouldechova2018case, Hurley18child}. For example, it is clearly impermissible that race has any direct effect on the decision made by the hotline screener, i.e., that all else being held fixed, members from one group have a higher probability of being surveilled by the agency. However, it is perhaps permissible that race has an indirect effect via some mediated pathway, e.g., if race is associated with some behaviors or features which themselves ought to be taken into consideration by hotline staffers, because they are predictive of abuse. If that's true, then $S \rightarrow A$ would be labeled an impermissible pathway whereas $S \rightarrow M \rightarrow A$ (for some $M$) would be permissible. Similarly, it would be unacceptable if race had an effect on whether children are separated from their families; arguably both the direct pathway $S \rightarrow Y_1$ and indirect pathway though hotline decisions $S \rightarrow A \rightarrow Y_1$ should be considered impermissible. Rather than defend any particular choice of path-specific constraints, we note that the framework outlined in \citet{NabiShpitser18Fair} can flexibly accommodate any set of given constraints, as long as the PSEs are identifiable from the observed distribution.

\subsection{Inference in a nearby ``fair world''}

We now describe the specifics of the proposal.
We assume the data is generated according to some (known) causal model, with observed data distribution $p(\cdot)$, and that we can characterize the fair world by a fair distribution $p^*(\cdot)$ where some set of pre-specified PSEs are constrained to be zero, or within a tolerance range. 
Without loss of generality we can assume the utility variable $Y$ is some deterministic function of $Y_1$ and $Y_2$ (i.e., $Y \equiv u(Y_1,Y_2)$) and thus use $Y$ in place of $Y_1$ and $Y_2$ in what follows.
Then $Z = (Y,X,S,M,A)$ in our child welfare example. For the purposes of illustration, assume the following two PSEs are impermissible:
PSE$^{sa}$, corresponding to the direct effect of $S$ on $A$ and defined as $\E[A(s,M(s'))] - \E[A(s')]$, and
PSE$^{sy}$, corresponding to the effect of $S$ on $Y$ along the edge $S \to Y$, and the path $S \to A \to Y$ and defined as $\E[Y(s, A(s,M(s')), M(s'))] - \E[Y(s')]$.

If the PSEs are identified under the considered causal model, they can be written as functions of the observed distribution. For example, the unfair PSE of the sensitive feature $S$ on outcome $Y$ in our child welfare example may be written as a functional PSE$^{sy} = g_1(Z) \equiv g_1(p(Y, X, S, M, A))$. Similarly the unfair PSE of $S$ on $A$ is PSE$^{sa}$ $= g_2(Z) \equiv g_2(p(Y, X, S, M, A))$. Generally, given a set of identified PSEs $g_j(Z)$ $\forall j \in \{ 1,..., J \}$ and corresponding tolerated lower/upper bounds $\epsilon_j^-, \epsilon_j^+$, the fair distribution $p^*(Z)$ is defined:
{\small
	\begin{align}
	p^*(Z) &\equiv \arg \min_{q} \hspace{0.1cm} D_{KL}(p||q) \nonumber \\
	\text{ subject to} \hspace{0.2cm} & \epsilon_j^- \leq g_j(Z) \leq \epsilon_j^+, \hspace{0.2cm} \forall j \in \{ 1,...,J \},
	\label{eqn:p-star}
	\end{align}
}%
where $D_{KL}$ is the KL-divergence and $J$ is the number of constraints.\footnote{Note that in our examples $J$ will typically be $K+1$, i.e., one constraint for the $S$ to $Y$ paths and one constraint for each set of paths from $S$ to $A_k$. We allow for $J$ constraints in general to accommodate more complex settings (e.g., where there are multiple sensitive features, multiple outcomes, or a different set of pathways are constrained).} In finite sample settings, \citet{NabiShpitser18Fair} propose solving the following constrained maximum likelihood problem:
{\small
\begin{align}
\widehat{\alpha} &= \arg \max_{\alpha} \hspace{0.1cm} {\cal L}(Z; {\alpha}) \nonumber \\
\text{ subject to} \hspace{0.2cm} & \epsilon_j^- \leq \widehat{g}_j(Z) \leq \epsilon_j^+, \hspace{0.2cm} \forall j \in \{ 1,...,J \},
 \label{eqn:c-mle}
\end{align}
}%
where $\widehat{g}_j(Z)$ are estimators for the chosen PSEs and ${\cal L}(Z; \alpha)$ is the likelihood function. The most relevant bounds in practice are $\epsilon_j^- =  \epsilon_j^+ = 0$.

Given an approximation of $p^*$ learned in this way, \citet{NabiShpitser18Fair} transform regression problems originally defined on $p$ into regression problems on $p^*$.
In other words, instead of learning a regression function $\mathbb{E}[Y \mid M, A, S, X]$ on the observed data distribution $p(Z)$, they approximate the fair distribution $p^*(Z)$ by constrained maximum likelihood and classify new instances using the constrained model. As we discuss in more detail later, \citet{NabiShpitser18Fair} choose to average over certain variables to accomodate the fact that new instances are generated from $p$ rather than $p^*$.


\subsection{Fair decision-making}

In the sequential decision setting, there are multiple complications.
In particular, we aim to learn high-quality policies while simultaneously making sure that the joint distribution induced by the 
policy satisfies our fairness criteria, potentially involving constraints on multiple causal pathways.  This problem must be solved in settings where distributions of some variables, such as outcomes, are not under the policy-maker's control.  Finally, we must show that if the learned policy is adapted to new instances (drawn from the original observed distribution) in the right way, then these new instances combined with the learned policy, constrained variables, and variables outside our control, together form a joint distribution where our fairness criteria remain satisfied.

Consider a $K$-stage decision problem given by a DAG where every vertex pair is connected, and with vertices in a topological order $X, S, M, A_1, Y_1, \ldots, A_K, Y_K$. See Fig.\ 1(c). Note that the setting where $S$ can be assumed exogenous is a special case of this model with missing edge between $X$ and $S$. Though we only assume a single set of permissible mediators $M$ here, at the expense of some added cumbersome notation all of the following can be extended to the case where there are distinct sets of mediators $M_1, \ldots, M_K$ preceding every decision point. (We extend the results below to that setting in the Supplement.)
We will consider the following PSEs as inadmissible: PSE$^{sy}$, representing the effect of $S$ on $Y$ along all paths \emph{other than} the paths of the form $S \to M \to \ldots \to Y$; and PSE$^{sa_k}$, representing the effect of $S$ on $A_k$ along all paths \emph{other than} the paths of the form $S \to M \to \ldots \to A_k$. That is, we consider \emph{only} pathways connecting $S$ and $A_k$ or $Y$ through the allowed mediators $M$ to be fair. 
In this model, these PSEs are identified by \cite{shpitser13cogsci}:
{\small
	\begin{align*}
	\text{PSE}^{sy} 
	&= \E[Y(s,M(s'))] - \E[Y(s')] \\ 
	= \sum_{X,M} &\{ \E[Y | s,M,X] - \E[Y | s',M,X] \} p(M | s',X) p(X)\\ \\
	\text{PSE}^{sa_k} 
	&= \E[A_k(s,M(s'))] - \E[A_k(s')] \\
	= \sum_{X,M} &\{ \E[ A_k | s,M,X] - \E[ A_k | s',M,X] \} p(M | s',X) p(X)
	\end{align*}
}%
Numerous approaches for estimating and constraining these identified PSEs are possible. In this paper, we restrict our attention to semiparametric estimators, which model only a part of the likelihood function while leaving the rest completely unrestricted.  Estimators of this sort share some advantages with parametric methods (e.g., often being uniformly consistent at favorable rates), but do not require specification of the full probability model. Specifically, we use estimators based on the following result:
\begin{thm}
	Assume $S$ is binary. Under the causal model above, the following are consistent estimators of PSE$^{sy}$ and PSE$^{sa_k}$, assuming all models are correctly specified:
	{\small
		\begin{align}
		&\widehat{g}^{sy}(Z) = \\ \nonumber 
		&\frac{1}{N} \sum_{n=1}^{N} \Big\{ \frac{\mathbb{I}(S_n = s)}{p(S_n | X_n)} \ \frac{p(M_{n} | s', X_n)}{p(M_{n} | s, X_n)}
		- \frac{\mathbb{I}(S_n = s')}{p(S_n | X_n)} \Big\} \ Y_n \\ \nonumber \\
		&\widehat{g}^{sa_k}(Z) =\\ \nonumber 
		 &\frac{1}{N} \sum_{n=1}^{N} \Big\{ \frac{\mathbb{I}(S_n = s)}{p(S_n | X_n)} \ \frac{p(M_{n} | s', X_n)}{p(M_{n} | s, X_n)}
		- \frac{\mathbb{I}(S_n = s')}{p(S_n | X_n)} \Big\} \ A_{kn}
		\end{align}
	}
	\label{thm:est}
\end{thm}
These inverse probability weighted (IPW) estimators use models for $M$ and $S$. Thus, we can approximate $p^*$ by constraining only the $M$ and $S$ models, i.e., obtaining estimates $\hat{\alpha}_{m}$ and $\hat{\alpha}_s$ of the parameters $\alpha_{m}$ and $\alpha_s$ in $p^*(M|S,X;\alpha_{m})$ and $p^*(S|X;\alpha_s)$ by solving (\ref{eqn:c-mle}). The outcomes $Y_k$ and decisions $A_k$ are left unconstrained. This is subtle and important, since it enables us to choose our optimal decision rules $f^*_A$ without restriction of the policy space and allows 
the mechanism determining outcomes $Y_k$ (based on decisions $A_k$ and history $H_k$) to remain 
outside the control of the policy-maker. Consequently, we can show that implementing this procedure guarantees that the joint distribution over all variables $Z$ induced by 1) the constrained $M$ and $S$ models, 2) the conditional distributions for $A_k$ given $H_k$ implied by the optimal policy choice, and 3) \emph{any} choice of $p(Y_k | A_k,H_k)$ will (at the population-level) satisfy the specified fairness constraints. We prove the following result in the Supplement:
\begin{thm}
	Consider the K-stage decision problem described by the DAG in Fig.~\ref{fig:triangle}(c). Let $p^*(M|S,X; \alpha_{m})$ and $p^*(S|X; \alpha_s)$ be the constrained models chosen to satisfy PSE$^{sy} = 0$ and PSE$^{sa_k} = 0$. Let $\tilde{p}(Z)$ be the joint distribution induced by $p^*(M|S,X; \alpha_{m})$ and $p^*(S|X;\alpha_s)$, and where all other distributions in the factorization are unrestricted. That is,
	{\small
	\begin{align*}
	\tilde{p}(Z) \equiv p(X)p^*(S|X;\alpha_s) &p^*(M|S,X;\alpha_m)\\
	 &\times \prod_{k=1}^K p(A_k|H_k) p(Y_k|A_k,H_k).\\
	\end{align*}
	}%
	Then the functionals PSE$^{sy}$ and PSE$^{sa_i}$ taken w.r.t.\ $\tilde{p}(Z)$ are also zero.
\end{thm}
This theorem implies that any approach for learning policies based on $\tilde{p}(Z)$ addresses both retrospective bias (since the fairness criterion violation present in $p(Z)$ is absent in $\tilde{p}(Z)$) and prospective bias (since the criterion holds in $\tilde{p}(Z)$ for any choice of policy on $A_k$ inducing $p(A_k | H_k)$).
As we discuss in detail in the next section, modified policy learning based on $\tilde{p}(Z)$
requires special treatment of the constrained variables $S$ and $M$. New instances (e.g., new calls to the child protection hotline) will be drawn from the unfair distribution $p$, not $\tilde{p}$. So, the enacted policy cannot use empirically observed values of $S$ or $M$. In what follows, our approach is to either average over $S$ and $M$ (following \citet{NabiShpitser18Fair}), or resample observations of $S$ and $M$ from the constrained models.

\section{Estimation of optimal policies in the fair world} 
\label{sec:estfairpolicy}
In the following, we describe several strategies for learning optimal policies, and our modifications to these strategies based on the above fairness considerations.



\subsection{Q-learning} 

In Q-learning, the optimal policy is chosen to optimize a sequence of counterfactual expectations called Q-functions.
These are defined recursively in terms of value functions $V_k(\cdot)$ as follows:
{\small
\begin{align}
Q_K(H_K, A_K) &= \E[Y_K(A_K) \mid H_K],  \nonumber \\
 V_K(H_K) &= \max_{a_K} Q_K(H_K, a_K), 
 \label{eq:QK}
\end{align}
}
and for $k = K-1, \ldots, 1$  
{\small
\begin{align}
Q_k(H_k, A_k) &= \E[V_{k+1}(H_{k+1},A_k) \mid H_k], \nonumber \\
 V_k(H_k)  &=  \max_{a_k} Q_k(H_k, a_k).
 \label{eq:Qi}
\end{align}
}%
Assuming $Q_k(H_k, A_k)$ is parameterized by $\beta_k$, the optimal policy at each stage may be easily derived from Q-functions as
$f^*_{A_k}(H_k) = \argmax_{a_k} {Q}_k(H_k, a_k; \widehat{\beta}_k)$.
Q-functions are recursively defined regression models where outcomes are value functions, and features are histories up to the current decision point.  Thus, parameters $\beta_k$ $(k=1,\ldots,K)$ of all Q-functions may be learned recursively by maximum likelihood methods applied to regression at stage $k$, given that the value function at stage $k+1$ was already computed for every row. See \citet{Moodie13DTR} for more details.

Note that at each stage $k$, the identity
$Q_k(H_k, A_k) = \E[V_{k+1}(H_{k+1},A_k) \mid H_k] = \E[V_{k+1}(H_{k+1}) \mid A_k, H_k]$ only holds under our causal model if the \emph{entire past} $H_k$ is conditioned on.  In particular, $\E[V_{k+1}(H_{k+1},A_k) \mid H_k \setminus \{ M, S \} ] \neq \E[V_{k+1}(H_{k+1}) \mid A_k, H_k \setminus \{M, S \} ]$.  To see a simple example of this, note that $Y_K(a_1)$ is not independent of $A_1$ conditional on just $X$ in Fig.~\ref{fig:triangle}(c), due to the presence of the path $Y_K \gets M \to A_1$; however the indepenence does hold conditional on the entire $H_1 = \{X, S, M\}$ \cite{thomas13swig}.

In a fair policy learning setting, though $\{M, S\}$ may be in $H_k$, we cannot condition on values of ${M, S}$ to learn fair policies since these values were drawn from $p$ rather than $p^*$. There are multiple ways of addressing this issue. One approach is to modify the procedure to obtain optimal policies that condition on all history \emph{other than} $\{M, S\}$. We first learn $Q_k$s using (\ref{eq:QK}) and (\ref{eq:Qi}). We then provide the following modified definition of Q-functions defined directly on $p^*$: 
{\small
\begin{align*} 
& Q^{*}_k(H_k \setminus \{M, S\}, A_k; \beta_k) = \\ 
& \frac{1}{Z} \ \sum_{m, s} Q_k(H_k, A_k; \beta_k) \ \prod_{i = 1}^{k} p(A_i | H_i \setminus \{M, S\}, m, s) .  \\
& \prod_{i = 2}^{k - 1}  p(M_i | A_i, H_i \setminus \{M, S\}, m, s) \ p^*(m, s | X),
\end{align*}
}%
for $k = K, \ldots, 1$, 
{\small
\begin{align*}      
Z &= \sum_{m, s} p^*(m, s | X) \ \prod_{i = 1}^{k} p(A_i | H_i \setminus \{M, S\}, m, s)  \\
& \prod_{i = 2}^{k - 1}  p(M_i | A_i, H_i \setminus \{M, S\}, m, s).   
\end{align*}    
}%
The optimal fair policy at each stage is then derived from $Q^{*}$-functions as $f^*_{A_k}(H_k) = \argmax_{a_k} {Q^{*}}_k(H_k \setminus \{M, S\}, a_k; \widehat{\beta}_k)$.

As an alternative approach, we can compute the original $Q$-functions defined in  (\ref{eq:QK}) and (\ref{eq:Qi}) with respect to $p^{*}(Z)$ by 
ignoring the observed values $M_n$ and $S_n$ for the $n$th individual and replacing them with samples drawn from $p^{*}(M|S,X;\alpha_m)$ and $p^{*}(S|X;\alpha_s)$.
Then, in (\ref{eq:QK}) and (\ref{eq:Qi}), the history at the $k$th stage, $H_k$, gets replaced with $H^{*}_k = \{H_k \setminus \{M, S\}, M^{*},  S^{*} \}$.

\subsection{Value search} 
It may be of interest to estimate the optimal policy within a restricted class ${\cal F}$.
One approach to learning the optimal policy within ${\cal F}$ is to directly search for the optimal ${ f}^{*,{\cal F}}_{ A} \equiv \argmax_{{f}_{A} \in {\cal F}} \E[Y({f}_{A})]$, which is known as \textit{value search}. 

The expected response to an arbitrary policy $\phi = \E[Y({f}_{A})]$, for ${f_A} \in {\cal F}$ can be estimated in a number of ways. 
Often $\widehat{\phi}$ takes the form of a solution to some estimating equation $\E[ h(\phi) ] = 0$ solved empirically given samples from $p(Z)$.  
A simple estimator for $\phi$ that uses only the treatment assignment model $\pi(H_k; \psi) \equiv p(A_k=1 | H_k)$ is the IPW estimator that solves the following estimating equation:
{\small
\begin{align}
\label{eqn:ipw}  
\E\left[  \prod_{k = 1}^{K} \left\{ C_{f_{A_k}} / \pi_{f_{A_k}}(H_k; \widehat{\psi})  \right\} \times Y  - \phi \right] = 0, 
\end{align}
}%
where $\small C_{f_{A_k}} \equiv \mathbb{I}(A_k = f_{A_k}(H_k))$,
$\pi_{f_{A_k}}(H_k;\psi) \equiv \pi(H_k;\psi) f_{A_k}(H_k) + (1-\pi(H_k;\psi)) (1 - f_{A_k}(H_k))$,
the expectation is evaluated empirically, and $\widehat{\psi}$ is fit by maximum likelihood.

Finding fair policies via value search involves solving the same problem with respect to $p^*(Z)$ instead. Given known models $p^*(M | S,X; \alpha_m)$ and $p^*(S | X; \alpha_s)$, we may consider two approaches. The first one involves solving a modified estimating equation of the form
{\small
\begin{align*}
&\E^*[ h(\phi) ] \equiv\\
&\E\left[ \sum_{m, s} \E[ h(\phi) | M,S,X ] p^*(M | S,X; \alpha_m) p^*(S | X; \alpha_s) \right] = 0
\end{align*}
}%
with respect to $p^*(Z \setminus \{M, S \})$. 
The alternative is to solve the original estimating equation $\E[ h(\phi) ] = 0$ with respect to $p^*(Z)$ by replacing observed values $M_n$ and $S_n$ for the $n$th individual with sampled values $M^*_n$ and $S^*_n$ drawn from $p^{*}(M|S,X;\alpha_m)$ and $p^{*}(S|X;\alpha_s)$.
In both approaches, the optimal fair policy at each stage is then derived by replacing the history at the $k$th stage, $H_k$, with $H^*_k = \left\{H_k \setminus \{M, S\}, M^*, S^* \right\}$.
Given constrained models $p^*(M | S,X; \alpha_m)$, and $p^*(S | X; \alpha_s)$ representing $p^*(Z)$, we can perform value search by solving the given estimating equation empirically on a dataset where every row $x_n, s_n, m_n$ in the data is replaced with $I$ rows $x_n, s^*_{ni}, m^*_{ni}$ for $i = 1, \ldots I$, 
with $m^*_{ni}$ and $s^*_{ni}$ drawn from $p^*(M | S, x_n; \alpha_m)$ and $p^*(S | x_n; \alpha_s)$, respectively. 

\subsection{G-estimation} 
A third method for estimating policies is to directly model the counterfactual contrasts known as \emph{optimal blip-to-zero functions} and then learn these functions by a method called g-estimation \citep{robins04SNMM}.  In the interest of space, we defer a full description of blip-to-zero functions and g-estimation to the Supplement, where we also present some results for our implementation of fair g-estimation.


\subsection{Tradeoffs and treatment of constrained variables}



We've proposed to constrain the $M$ and $S$ models to satisfy given fairness constraints. Since empirically observed values of $M$ and $S$ are sampled from $p$ rather than $p^*$ (or $\tilde{p}$), our approach requires resampling or averaging over these features. The choice of models to constrain involves a tradeoff.  The more models are constrained, the closer the KL distance between $p$ and $p^*$, but the more features have to be resampled or averaged out; that is, some information on new instances is ``lost.'' Alternative approaches may constrain fewer or different models in the likelihood (for example, we could have elected to constrain the $Y$ model instead of $S$). However, the benefit of our approach here is that we can guarantee, with outcomes $Y$ outside the policy-maker's control, that the induced joint distribution will satisfy the given fairness constraints (by Theorem 2), whereas alternative procedures which aim to avoid averaging or resampling will typically have no such guarantees. Another alternative that avoids averaging over variables altogether is to consider likelihood parameterizations where the absence of a given PSE directly corresponds to setting some variation-independent likelihood parameter for the $Y$ model to zero (cf.\ \citet{chiappa2018path}). While such a parameterization is possible for linear structural equation models, it is an open problem in general for arbitrary PSEs and nonlinear settings.
Developing novel, general-purpose alternatives that transfer observed distributions to their ``fair versions,'' while avoiding resampling and averaging, is an open problem left to future work.

\section{Experiments}
\label{sec:exp}

\subsection*{Synthetic data}
 
We generated synthetic data for a two-stage decision problem according to the causal model shown in Fig.~\ref{fig:triangle}(c) ($K = 2$), where all variables are binary except for the continuous response utility $Y \equiv Y_2$. Details on the specific models used are reported in the Supplement. We generated a dataset of size $5,000$, with $100$ bootstrap replications, where the sensitive variable $S$ is randomly assigned and where $S$ is chosen to be an informative covariate in estimating $Y$. 

We use estimators in Theorem \ref{thm:est} to compute PSE$^{sy}$, PSE$^{sa_1}$, and PSE$^{sa_2}$ which entail using $M$ and $S$ models.  In this setting, the PSE$^{sy}$ is $1.918$ (on the mean scale) and is restricted to lie between $-0.1$ and $0.1$. The PSE$^{sa_1}$ is $0.718$, and PSE$^{sa_2}$ is $0.921$ (on the odds ratio scale) and both are restricted to lie between $0.95$ and $1.05$. We only constrain $M$ and $S$ models to approximate $p^*$ and fit these two models by maximizing the constrained likelihood using the R package \texttt{nloptr}. The parameters in all other models were estimated by maximizing the likelihood. 

Optimal fair polices along with optimal (unfair) policies were estimated using the two techniques described in Section \ref{sec:estfairpolicy} (where we used the ``averaging'' approach in both cases). We evaluated the performance of both techniques by comparing the population-level response under fair policies versus unfair policies.
One would expect the unfair policies to lead to higher expected outcomes compared to fair policies since satisfying fairness constraints requires sacrificing some policy effectiveness. The expected outcomes under unfair polices obtained from Q-learning and value search were  $7.219 \!\pm\! 0.005$ and $7.622 \!\pm\! 0.265$, respectively. The values dropped to $6.104 \!\pm\! 0.006$ and $6.272 \!\pm\! 0.133$ under fair polices, as expected. 
In addition, both fair and unfair optimal polices had higher expected outcomes than the observed population-level outcome, using both methods. In our simulations, the population outcome under observed policies was  $4.82 \!\pm\! 0.007$. Some additional results are reported in the Supplement. 

%

\subsection*{Application to the COMPAS dataset}
COMPAS is a criminal justice risk assessment tool created by the company Northpointe that has been used across the US to determine whether to release or detain a defendant before their trial. Each pretrial defendant receives several COMPAS scores based on factors including but not limited to demographics, criminal history, family history, and social status. Among these scores, we are primarily interested in the ``risk of recidivism." We use the data made available by Propublica and described in \citet{angwin16machine}.  
The COMPAS risk score for each defendant ranges from 1 to 10, with 10 being the highest risk. In addition to this score ($A$), the data also includes records on defendant’s age ($X_1 \in X$), gender ($X_2 \in X$), race ($S$), prior convictions ($M$), and whether or not recidivism occurred in a span of two years ($R$). We limited our attention to the cohort consisting of African-Americans and Caucasians, and to individuals who either had not been arrested for a new offense or who had recidivated within two years. Our sample size is 5278. All variables were binarized including the COMPAS score, which we treat as an indicator of a binary decision to incarcerate versus release (pretrial) ``high risk'' individuals, i.e., we assume those with score $\geq 7$ were incarcerated. In this data, 28.9\% of individuals had scores $\geq 7$.

Since the data does not include any variable that corresponds to utility, and there is no uncontroversial definition of what function one should optimize, we define a heuristic utility function from the data as follows. We assume there is some (social, economic, and human) cost, i.e., negative utility, associated with incarceration (deciding $A=1$), and that there is some cost to releasing individuals who go on to reoffend (i.e., for whom $A=0$ and $R=1$). Also, there is positive utility associated with releasing individuals who do not go on to recidivate (i.e., for whom $A=0$ and $R=0$). A crucial feature of any realistic utility function is how to balance these relative costs, e.g., how much (if any) ``worse'' it is to release an individual who goes on to reoffend than to incarcerate them. To model these considerations we define utility $Y \equiv (1-A) \times \{ -\theta R + (1-R) \} - A$.
The utility function is thus parameterized by $\theta$, which quantifies how much ``worse'' is the case where individuals are released and reoffend as compared with the other two possibilities which are treated symmetrically. 
We emphasize that this utility function is a heuristic we use to illustrate our optimal policy learning method, and that a realistic utility function would be much more complicated (possibly depending also on factors not recorded in the available data). 

We apply our proposed Q-learning procedure to optimize $\E[Y]$, assuming $K=1$ and exogenous $S$. The fair policy constrains the $S \rightarrow A$ and $S \rightarrow Y$ pathways. We describe details of our implementation as well as additional results in the Supplement. The proportion of individuals incarcerated ($A=1$) is a function of $\theta$, which we plot in Fig.\ 2 stratified by racial group. See the Supplement for results on \emph{overall} incerceration rates, which also vary among the policies.
The region of particular interest is between $\theta=2$ and $3$, where fair and unrestricted optimal policies differ and both recommend lower-than-observed overall incarceration rates (see Supplement). For most $\theta$ values, the fair policy recommends a decision rule which narrows the racial gap in incarceration rates as compared with the unrestricted policy, though does not eliminate this gap entirely. (Constraining the causal effects of race through mediator $M$ would go further in eliminating this gap.) In regions where $\theta >3$, both optimal policies in fact recommend higher-than-observed overall incarceration rates 
but a narrower racial gap, particularly for the fair policy.
Comparing fair and unconstrained policy learning on this data serves to simultaneously illustrate how the proposed methods can be applied to real problems and how the choice of utility function is not innocuous.
\begin{figure}[t!]
\begin{center}
\includegraphics[scale=.32]{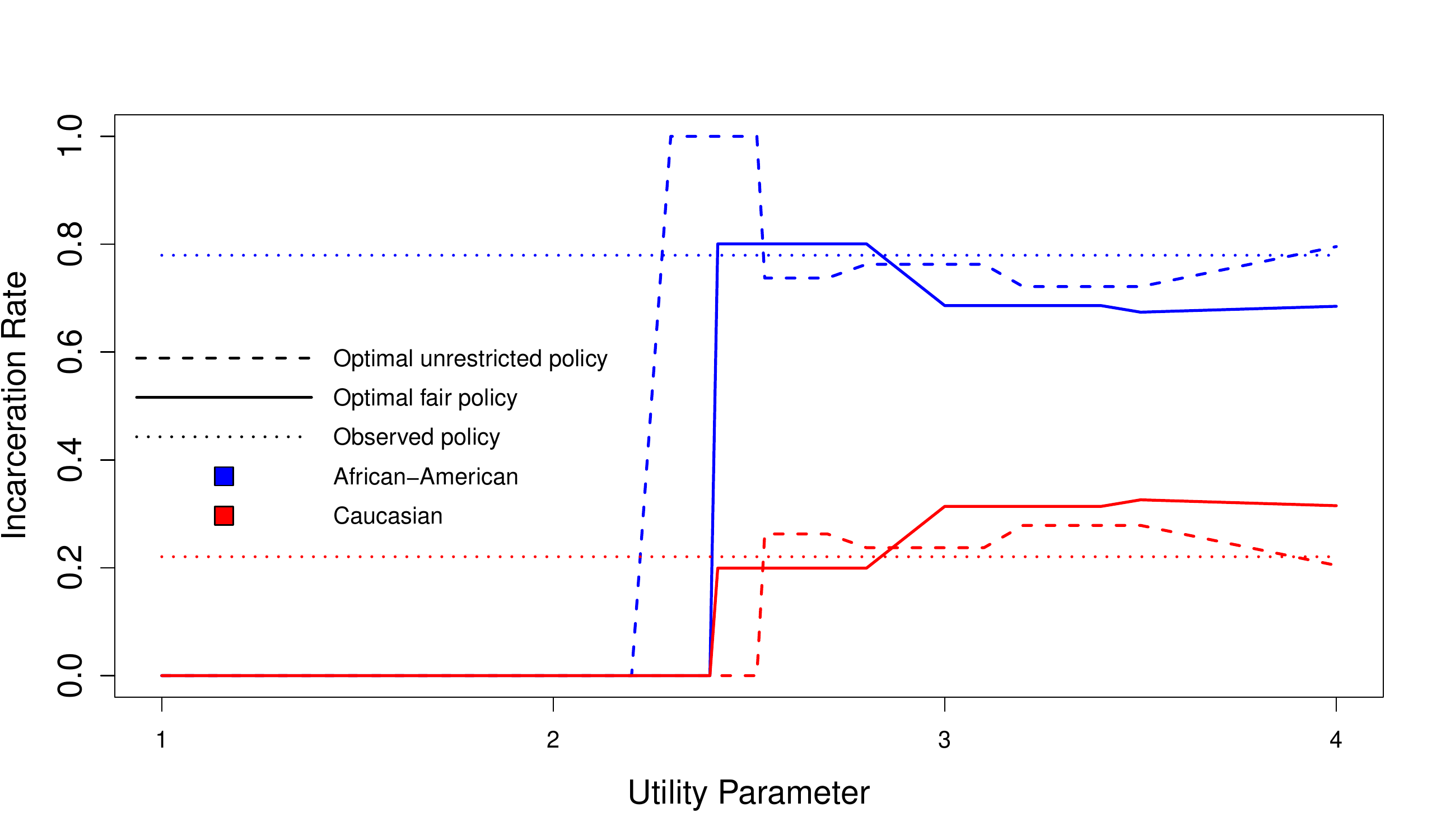}
\end{center}
\caption{Group-level incarceration rates 
	for the COMPAS data as a function of the utility parameter $\theta$.}
\vspace{-4.7mm}
\label{sim}
\end{figure}

\section{Conclusion}
\label{sec:conc}

We have extended a formalization of algorithmic fairness from \citet{NabiShpitser18Fair} to the setting of learning optimal policies under fairness constraints.
We show how to constrain a set of statistical models and learn a policy such that
subsequent decision making given new observations from the ``unfair world'' induces high-quality outcomes while satisfying the specified fairness constraints in the induced joint distribution.  
In this sense, our approach can be said to ``break the cycle of injustice'' in decision-making.
We investigated the performance of our proposals on synthetic and real data, where in the latter case we have supplemented the data with a heuristic utility function. 
In future work, we hope to develop and implement more sophisticated constrained optimization methods, to use information as efficiently as possible while satisfying the desired theoretical guarantee, and to explore nonparametric techniques for complex settings where the likelihood is not known.


\clearpage

\section*{Acknowledgments}
This project is sponsored in part by the National Institutes of Health grant R01 AI127271-01 A1 and the Office of Naval Research grant N00014-18-1-2760. 

{\small

\bibliography{references}
\bibliographystyle{icml2019}
}
\end{document}


%

\twocolumn[
\icmltitle{Supplemental Appendix for ``Learning Optimal Fair Policies''}



\icmlsetsymbol{equal}{*}

\begin{icmlauthorlist}
	\icmlauthor{Razieh Nabi}{to}
	\icmlauthor{Daniel Malinsky}{to}
	\icmlauthor{Ilya Shpitser}{to}
\end{icmlauthorlist}

\icmlaffiliation{to}{Department of Computer Science, Johns Hopkins University, Baltimore, MD, USA}

\icmlcorrespondingauthor{Razieh Nabi}{rnabi@jhu.edu}

\icmlkeywords{Fair Policies, Policy Learning, Causal Inference, Mediation Analysis, Algorithmic Fairness}

\vskip 0.3in
]

\printAffiliationsAndNotice{}  

\section*{Appendix A: G-estimation}
G-estimation applies to structural nested models, which directly model the counterfactual deviations in outcome from a reference treatment value (which we take to be $A=0$) conditional on history, assuming all future decisions are already optimal. Specifically, for each decision point $k$ we posit a \emph{structural nested mean
	model (SNMM)} parameterized by $\psi$ as follows: 
{\small
\begin{align*}
\gamma_k(H_k, &a_k; \psi) =\\
&\E[ Y(\bar{a}_{k-1}, a_k, f^*_{\underline{A}_{k+1}}) - Y(\bar{a}_{k-1}, a_k=0, f^*_{\underline{A}_{k+1}}) \mid H_k ],  
\end{align*}
}%
where $\underline{A}_{k+1}$ represents all treatments administered from time $k+1$ onwards. In words, $\gamma_k$ is the contrast of the counterfactual mean (conditional on observed history $H_k$) where the past decisions are set to their observed values, the present decision is either $a_k$ or a reference decision $a_k = 0$, and all future decisions are made optimally, $f^*_{\underline{A}_{k+1}}$.  

Note that if the true $\gamma_k(H_k,a_k;\psi)$ were known, the optimal treatment policies are those that maximize this ``blip'' function at each stage: $f^*_{A_k} = \argmax_{a_k} \gamma_k(H_k, a_k; \psi)$.
In order to estimate $\psi$ using data, let 
{\small
	\begin{align}
	U({\bf \psi, \zeta(\psi), \alpha}) = \sum_{k=1}^{K} &\left\{ G_k({\bf \psi}) - \E\left[G_k({\bf \psi}) \mid H_k; 
	{\bf \zeta} \right] \right\} 
	\nonumber \\
	\times \ &\left\{ d_k(H_k, A_k) - E\left[d_k(H_k, A_k) \mid H_k; {\bf \alpha}\right] \right\}, 
	\label{eqn:G-estim}
	\end{align}
}%
where $d_k(H_k, A_k)$ is any function of $H_k$ and $A_k$ and $G_k({\bf \psi})$ is defined as 
{\small
	\begin{align*}
	Y  - \gamma_k(H_k, a_k; {\bf \psi}) + \sum_{i = k + 1}^{K} \left[ \gamma_i(H_i, a^*_i; {\bf \psi}) - \gamma_i(H_i, a_i; {\bf \psi}) \right],
	\end{align*}
}%
($a^*_i$ is the optimal decision at $i$th stage).   
Consistent estimators of ${\bf \psi}$ can be obtained solving the estimating equations $\E[U({\bf \psi, \zeta(\psi), \alpha})] = 0$, as shown in \citet{robins04SNMM}. 

Both of the modifications discussed for Q-learning and value search must be applied when learning fair optimal policies by g-estimation.  Specifically, we determine optimal polities not from the SNMM contrast
$\gamma_k(H_k, a_k; \psi) = \E[ Y(\bar{a}_{k-1}, a_k, f^*_{\underline{A}_{k+1}}) - Y(\bar{a}_{k-1}, a_k = 0, f^*_{\underline{A}_{k+1}})\mid H_k ]$ itself, but rather from a modified contrast
$\gamma^*_k(H_k \setminus M, a_k; \psi) = \sum_{m, s} \gamma_k(H_k, a_k; \psi) p^*(M | S,X) p^*(S | X) =
\E[ Y(\bar{a}_{k-1},a_k, f^*_{\underline{A}_{k+1}}) - Y(\bar{a}_{k-1}, a_k = 0, f^*_{\underline{A}_{k+1}})\mid H_k \setminus  \{M, S\} ]$ which does not use $M$ and $S$. This is analogous to removing $M$ and $S$ from the Q-functions defined in Section 4 and is done for the same reason: $M, S$ are drawn from $p(Z)$, not $p^*(Z)$. 

Second, the estimating equations for $\psi$ must use constrained models (in particular for $M$ and $S$), and must be empirically solved using observations only from $p^*(Z)$.  As was done with value search, we solve equation (\ref{eqn:G-estim}) empirically using a dataset where each row $x_n,s_n,m_n$ is replaced by $I$ rows of the form $x_n,s^*_{ni},m^*_{ni}$, $i = 1, \ldots, I$, with $s^*_{ni}$ and $m^*_{ni}$ drawn from $p^*(S | x_n; \alpha_s)$ and $p^*(M | x_n, S; \alpha_m)$, respectively.

\section*{Appendix B: Simulation details and additional results on synthetic data} 
Here we report the precise parameter settings used in our simulation studies. The following regression models were used in our simulation study of the two-stage decision problem:
{\small
	\begin{align*}
	X_1 &\sim |  \mathcal{N}(0, 1) | 
	\nonumber \\
	(X_2, X_3)  &\sim  \mathcal{N}(0, \text{diag}(2)) 
	\nonumber \\ 
	S  &\sim \text{Bernoulli}(p = 0.5) 
	\nonumber \\  
	\text{logit}(p(M = 1)) &\sim -1 + X_1 + X_2 + X_3 + S\\
										&+ 3SX_1 + SX_2 + SX_3
	\nonumber \\
	\text{logit}(p(A_1 = 1)) &\sim 1 - X_1 + X_2 + S + M - SX_1 + SX_2\\
											&+ MS - 3MX_1 + 0.5MX_2 
	\nonumber \\
	\text{logit}(p(Y_1 = 1)) &\sim -2 + X_1 + X_2 + S + M + A\\
											&+ SX_2 + MS + AS + AM 
	\nonumber \\
	\text{logit}(p(A_2 = 1)) &\sim 1 - X_1 + X_2  + M + A + W\\
											&+ S(1 - X_1 + X_2 + M - A) \\
											&- 3MX_1 + 0.5MX_2 - AX_1 - AX_2 
	\nonumber \\
	Y& = 2.5 + X_1 + X_2 + M + W + B\\
		&+ S(1+ X_1+  X_2 + M + A + W) \\ 
		&+ A(1 + M - 2W) + MW\\
		&+  B(-X_1 + 2X_2 - M) + WX_1 + \mathcal{N}(0, 1) 
	\end{align*}
}%
For this two-stage setting we estimated the optimal policies using Q-learning and value search. 
In value search, we considered restricted class of polices of the form $p(A_1 = 1 | X, S, M) = -1 + \alpha_x X + \alpha_s S + \alpha_m M +   \alpha_{sx} SX + \alpha_{sm} SM + \alpha_{mx} MX $, and $p(A_2 = 1 | X, S, M, A_1, Y_1) = -1 + \alpha_x X + \alpha_s S + \alpha_m M +  \alpha_a A + \alpha_{y_1} Y_1 + \alpha_{sx} SX + \alpha_{sm} SM + \alpha_{mx} MX  + \alpha_{as} AS + \alpha_{ax} AX$ where all $\alpha$s range from $-3$ to $3$ by $0.5$ increments and estimated the value of policies for each combination of $\alpha$s using equation (\ref{eqn:ipw}). 

A third method for estimating policies is to directly model the counterfactual contrasts known as \emph{optimal blip-to-zero functions} and then learn these functions by g-estimation \citep{robins04SNMM}; see Appendix A. We implemented our modified fair g-estimation for a single-stage decision problem and compared the results with Q-learning and value search. The results are provided in Table 1. The data generating process for the single-stage decision problem matches the causal model shown in Fig.~\ref{fig:triangle}(a) where  $X, S, M,$ and $A$ were generated the same way as described above. The outcome $Y$ was generated from a standard normal distribution with mean $-2 + X + S + M + A -3SX_2 + MS + AS + AM + AX_2 + AX_3$. 
We used estimators in Theorem \ref{thm:est} to compute PSE$^{sy}$ and PSE$^{sa}$ which require using $M$ and $S$ models. In this synthetic data, the PSE$^{sy}$ was $1.618$ (on the mean scale) and was restricted to lie between $-0.1$ and $0.1$. The PSE$^{sa}$ was $0.685$ (on the odds ratio scale) and was restricted to lie between $0.95$ and $1.05$. 


\begin{table}[t!]
	\label{tab:outcome}
	\centering
	\begin{tabular}{ c | c | c }
		& \small \textbf{Unfair Policy} & \small \textbf{Fair Policy}  \\  [0.1cm]
		\hline
		\small \textbf{Q-learning }  & $ 1.414 \!\pm\! 0.0056 $ &  $ 1.189 \!\pm\! 0.0059 $ \\  [0.1cm]
		\small \textbf{value search }  & $ 1.134 \!\pm\! 0.0245 $ &  $ 1.056 \!\pm\!  0.0299 $ \\ [0.1cm]
		\small \textbf{g-estimation }  & $ 1.375  \!\pm\! 0.0099  $ &  $ 1.312 \!\pm\! 0.0102 $ 
	\end{tabular}
	\caption{{\small Comparison of population outcomes $\E[Y]$ under policies learned by different methods. The value under the observed policy was $ 0.24 \!\pm\! 0.006 $. }}
\end{table}

\section*{Appendix C: Details and additional results on the COMPAS data experiment}
The regression models we used in the COMPAS data analysis were specified as follows:
{\small
\begin{align*}
\text{logit}(p(M=1)) &\sim X_1 + X_2 + S + SX_1 + SX_2\\
\text{logit}(p(A=1)) &\sim X_1 + X_2 + S + M + SX_1\\
  & \hspace{0.1cm} + SX_2 + MS + MX_1 + MX_2\\
Y &\sim X_1 + X_2 + S + M + A + SX_1 + SX_2 \\
 & \hspace{0.1cm} + AS + AM + MS + MX_1 + MX_2\\ 
 & \hspace{0.1cm} + AX_1 + AX_2 
\end{align*} 
}%
For estimating the PSEs which we constrain, we used the same IPW estimators described in the main paper and reproduced in the theorem below. We constrained the PSEs to lie between $-0.05$ and $0.05$ and $0.95$ and $1.05$, respectively.

In Fig.\ 3, we compare the overall incarceration rates recommended by the optimal fair and unconstrained policies on the COMPAS data, as a function of the utility parameter $\theta$. For low values of $\theta$ the incarceration rate is zero, and becomes higher as $\theta$ increases, but differentially for the fair and unconstrained optimal policies. The difference between the policies depends crucially on the utility function. For some values of the utility parameter, the unfair and fair policies coincide, but for other values we would expect significantly different overall incarceration rates as well as different disparities between racial groups (see result in the main paper).

In Fig.\ 4, we show the relative utility achieved by the optimal fair and unconstrained policies, as well as the utility of the observed decision pattern, as a function of $\theta$. As expected, choosing an optimal policy improves on the observed policy, with the unfair (unconstrained) choice being higher utility than the fair (constrained) choice; we sacrifice some optimality to satisfy the fairness constraints. However, the difference depends on the utility parameter and for a range of parameter values the fair and unfair policies are nearly the same in terms of optimality (even when they may disagree on the resulting incarceration rate, around $\theta=2.6$). The fair and unfair policies drift far apart in terms of utility around $\theta=3$, when the policies recommend an incarceration rate comparable to or higher than the observed rate.
\addtocounter{figure}{2}
\begin{figure}[t!] 
	\begin{center}
		\includegraphics[scale=.32]{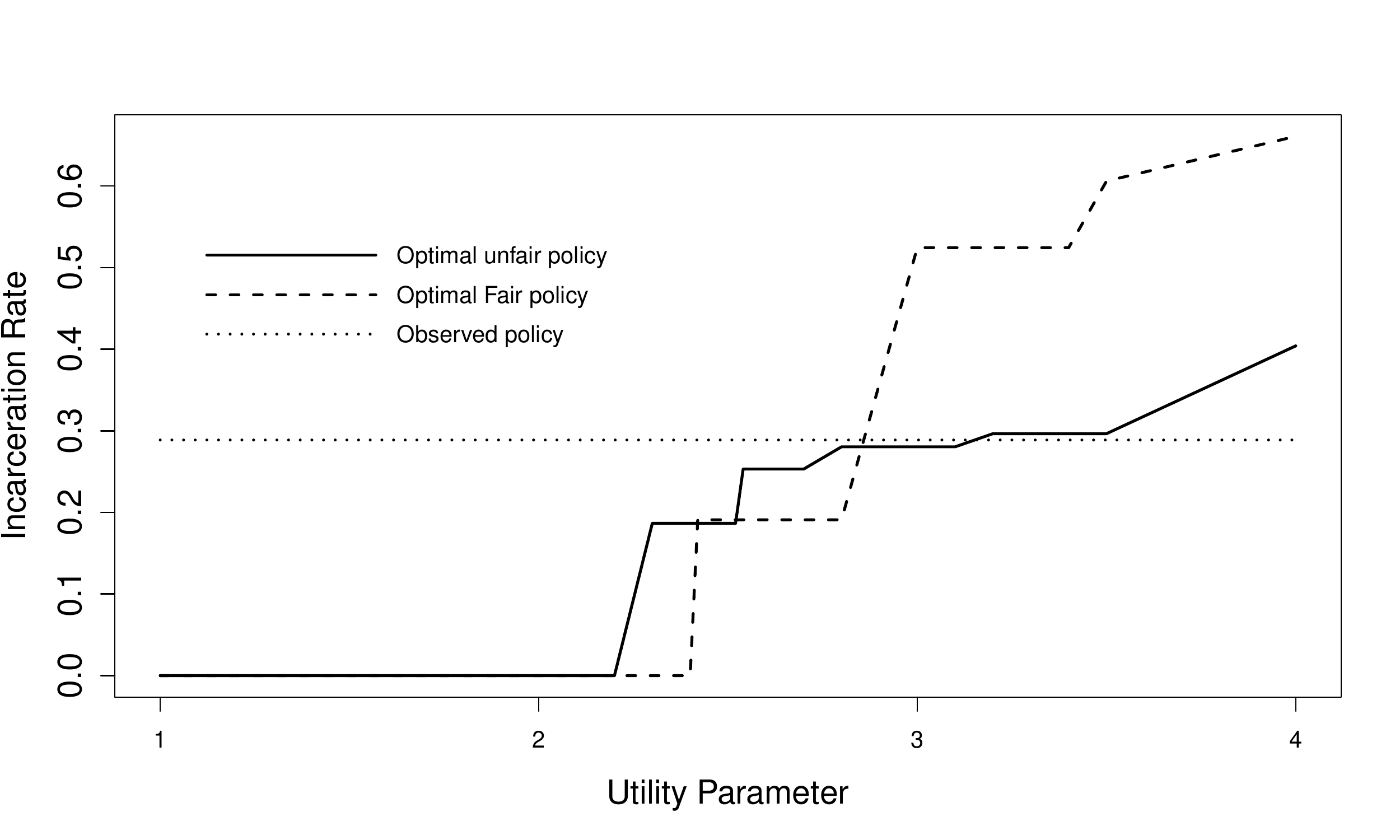}
	\end{center}
	\caption{Overall incarceration rates for the COMPAS data as a function of the utility parameter $\theta$.}
\end{figure}
\begin{figure}[t!] 
	\begin{center}
		\includegraphics[scale=.32]{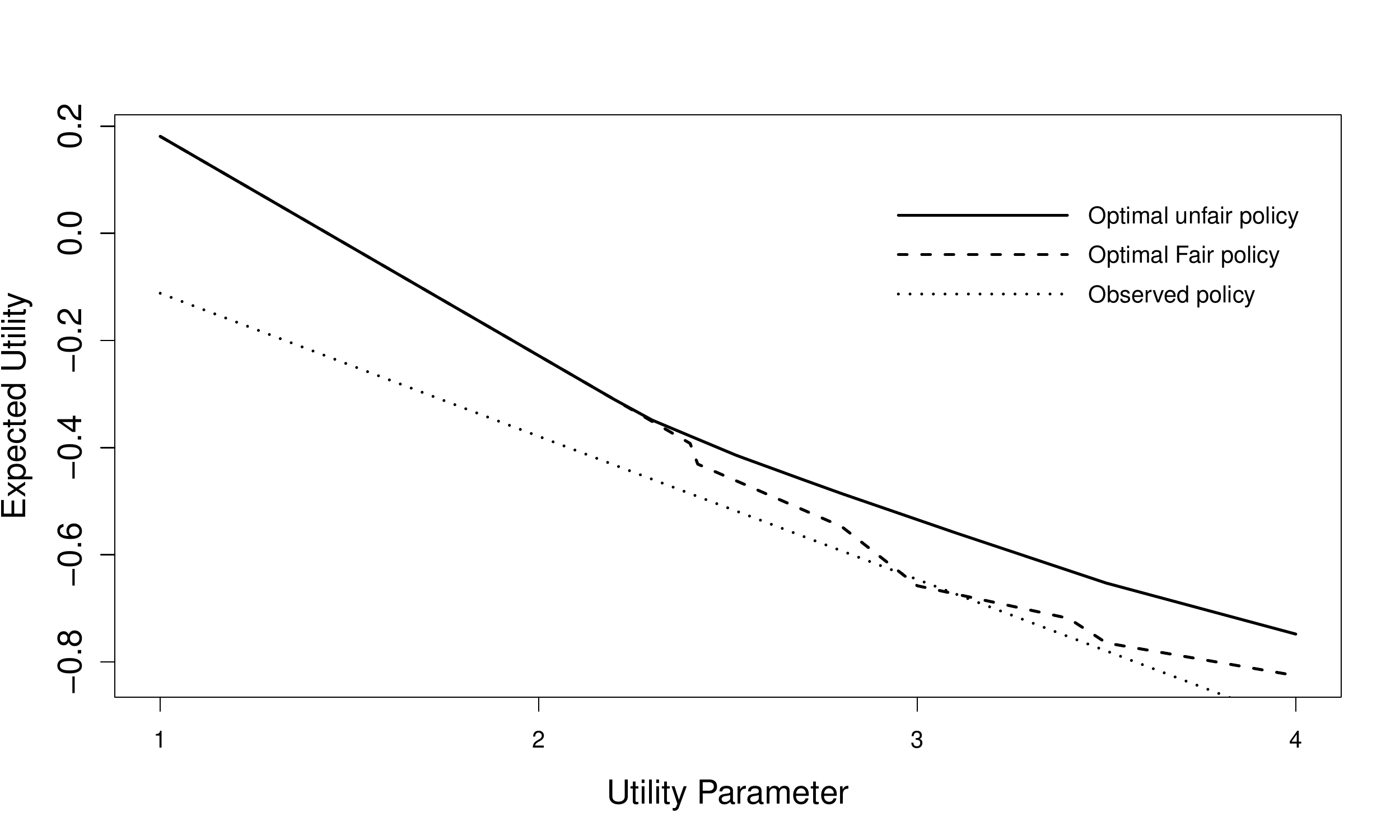}
	\end{center}
	\caption{The relative utility of policies for the COMPAS data as a function of the utility parameter $\theta$.}
\end{figure}

\newpage
\section*{Appendix D: Proofs}

\begin{thm}
	Assume $S$ is binary. Under the causal model above, the following are consistent estimators of PSE$^{sy}$ and PSE$^{sa_k}$, assuming all models are correctly specified:
	{\small
		\begin{align}
			&\widehat{g}^{sy}(Z) = \\ \nonumber 
			&\frac{1}{N} \sum_{n=1}^{N} \Big\{ \frac{\mathbb{I}(S_n = s)}{p(S_n | X_n)} \ \frac{p(M_{n} | s', X_n)}{p(M_{n} | s, X_n)}
			- \frac{\mathbb{I}(S_n = s')}{p(S_n | X_n)} \Big\} \ Y_n \\ \nonumber \\
			&\widehat{g}^{s{a}_k}(Z) =\\ \nonumber 
			&\frac{1}{N} \sum_{n=1}^{N} \Big\{ \frac{\mathbb{I}(S_n = s)}{p(S_n | X_n)} \ \frac{p(M_{n} | s', X_n)}{p(M_{n} | s, X_n)}
			- \frac{\mathbb{I}(S_n = s')}{p(S_n | X_n)} \Big\} \ A_{kn}
		\end{align}
	}%
\end{thm}
\begin{prf}
	The latent projection \cite{verma90equiv} of any $K$ stage DAG onto $X,S,M,A,Y$ suffices to identify and estimate the two path-specific effects in question, and this latent projection is the complete DAG with topological ordering $X,S,M,A,Y$.
	The consistency of the estimators above then follows directly from derivations in \cite{tchetgen12semi}.
	As an example, we have the following derivation for the first term of $g^{sy}(Z)$:
{\small
	\begin{align*}
	&
	\sum_{X,M} \E\left[
	Y| s, M, X
	\right]p(M|s',X)p(X)\\
	&= \sum_{X,M,A,Y} Y p(Y | s,M,A,X) p(A | s,M,X) p(M | s',X) p(X)\\
	&=
	\sum_{X,S,M,A,Y} \frac{\mathbb{I}(S = s)p(M | s', X)}{p(S | X) p(M | s, X)} Y 
	dp(Y,S,M,A,X)\\
	&= \E\left[
	\frac{\mathbb{I}(S = s)p(M | s', X)}{p(S | X) p(M | s, X)} Y 
	\right]\\
	\end{align*}
}%
	which is precisely the identifying functional for the first term of the PSE we are interested in. That the above estimator is consistent for this functional is a standard result.
\end{prf}

\begin{thm}
	Consider the K-stage decision problem described by the DAG in Fig 1c. Let $p^*(M|S,X; \alpha_{m})$ and $p^*(S|X; \alpha_s)$ be the constrained models chosen to satisfy PSE$^{sy} = 0$ and PSE$^{sa_k} = 0$. Let $\tilde{p}(Z)$ be the joint distribution induced by $p^*(M|S,X; \alpha_{m})$ and $p^*(S|X;\alpha_s)$, and where all other distributions in the factorization are unrestricted. That is,
	{\small
		\begin{align*}
		\tilde{p}(Z) \equiv p(X)p^*(S|X;\alpha_s) &p^*(M|S,X;\alpha_m)\\
		&\times \prod_{k=1}^K p(A_k|H_k) p(Y_k|A_k,H_k).\\
		\end{align*}
	}%
	Then the functionals PSE$^{sy}$ and PSE$^{sa_k}$ taken w.r.t.\ $\tilde{p}(Z)$ are also zero.
\end{thm}
\begin{prf}
Let $Y \equiv Y_K$.  Because $M$ preceeds all $A_k,Y_k$ for $k = 1, \ldots K$, it suffices to consider the latent projection
with only variables $X,S,M,A,Y$ without affecting identifiability considerations.  Then we have the following:
{\small
	\begin{align*}
	&\widetilde{\text{PSE}}^{sy} \\
	&= \tilde{\E}[Y(s,M(s'))] - \tilde{\E}[Y(s')] \\ 
	&= \sum_{X,M} \{ \tilde{\E}[Y | s,M,X] - \tilde{\E}[Y | s',M,X] \} p^*(M | s',X;{\alpha}_m) p(X) \\
	&= \sum_{X,M} \{ {\E}[Y | s,M,X] - {\E}[Y | s',M,X] \} p^*(M | s',X;{\alpha}_m) p(X) \\
	&= \sum_{X,M,Y} Y \{ p(Y | s,M,X) - p(Y | s',M,X) \} p^*(M | s',X;{\alpha}_m) p(X) \\
	&= \sum_{X,S,M,Y} Y \Big\{ \frac{\mathbb{I}(S = s)}{p^*(S | X;\alpha_s)} \ \frac{p^*(M | s', X;\alpha_m)}{p^*(M | s, X;\alpha_m)}
	- \frac{\mathbb{I}(S = s')}{p^*(S | X;\alpha_s)} \Big\}  \\
	& \times p(Y | M,S,X) p^*(M | S,X; \alpha_m) p^*(S | X; \alpha_s) p(X)\\
	&=0
	\end{align*}
}%
by choice of $p^*(M |S, X;\alpha_m)$ and $p^*(S | X;\alpha_s)$.
The proof is structurally the same for $\widetilde{\text{PSE}}^{sa_k}$.
\end{prf}

\section*{Appendix E: Modified results with multiple sets of mediators}
In the main paper, we discussed a $K$-stage decision problem with one set of permissible mediators, $M$. Here, we extend those results to the setting where we have multiple sets of mediators $M_1, \ldots, M_K$, i.e., a DAG with topological ordering $X, S, M_1, A_1, Y_1, \ldots, M_K, A_K, Y_K$. In this case, we consider the following paths impermissible: PSE$^{sy}$, representing the effect of $S$ on $Y$ along all paths \emph{other than} the paths of the form $S \to M_k \to \ldots \to Y$ ($\forall k$); and PSE$^{sa_k}$, representing the effect of $S$ on $A_k$ along all paths \emph{other than} the paths of the form $S \to M_j \to \ldots \to A_k$ ($\forall j \leq k$). That is, we consider \emph{only} pathways connecting $S$ and $A_k$ or $Y$ through the allowed mediators $M_1, \ldots, M_K$ to be fair. In this case, the PSEs are identified by a modification of the previous formula given in Section 3.2. 
{\small
\begin{align*}
	&\text{PSE}^{sy} = \E[Y(s,M_1(s'),\ldots,M_K(s'))] - \E[Y(s')] \\ 
	&= \sum_{x,\overline{m}_K,\overline{a}_{K-1},\overline{y}_{K-1},} \{ \E[Y | s,\overline{M}_K, \overline{A}_{K-1},\overline{Y}_{K-1},X]\\
	&- \E[Y | s', \overline{M}_K, \overline{A}_{K-1},\overline{Y}_{K-1},X] \} \prod_{k=1}^K p(M_k | s',\overline{A}_{k-1},\overline{Y}_{k-1},X)\\
	&\times \prod_{k=1}^{K-1} p(A_k | s, \overline{M}_k,\overline{A}_{k-1},\overline{Y}_{k},X) p(Y_k | s,\overline{M}_k,\overline{A}_{k},\overline{Y}_{k-1},X) p(X) 
	\\
	&\mbox{}
\end{align*}
}

\vspace{-1.5cm}
{\small
\begin{align*}
	&\text{PSE}^{sa_k} = \E[A_k(s,M_1(s'),\ldots,M_K(s'))] - \E[A_k(s')] \\ 
	&= \sum_{x,\overline{m}_k,\overline{a}_{k-1},\overline{y}_{k-1},} \{ \E[A_k | s,\overline{M}_k, \overline{A}_{k-1},\overline{Y}_{k-1},X]\\
	&- \E[A_k | s', \overline{M}_k, \overline{A}_{k-1},\overline{Y}_{k-1},X] \} \prod_{k=1}^K p(M_k | s',\overline{A}_{k-1},\overline{Y}_{k-1},X)\\
	&\times \prod_{j=1}^{k-1} p(A_j | s, \overline{M}_j,\overline{A}_{j-1},\overline{Y}_{j},X) p(Y_j | s,\overline{M}_j,\overline{A}_{j},\overline{Y}_{j-1},X) p(X)
\end{align*}
}%
With these definitions, we can replace the estimators in Theorem 1 with:
{\small
	\begin{align*}
	&\widehat{g}^{sy}(Z) = \\ \nonumber 
	&\frac{1}{N} \sum_{n=1}^{N} \Big\{ \frac{\mathbb{I}(S_n = s)}{p(S_n | X_n)} \ \prod_{k=1}^K \frac{p(M_{k,n} | s', \overline{A}_{k-1,n},\overline{Y}_{k-1,n}, X_n)}{p(M_{k,n} | s, \overline{A}_{k-1,n},\overline{Y}_{k-1,n}, X_n)}\\
	&- \frac{\mathbb{I}(S_n = s')}{p(S_n | X_n)} \Big\} \ Y_n 
	\\  \\
	&\widehat{g}^{sa_k}(Z) =\\ \nonumber 
	&\frac{1}{N} \sum_{n=1}^{N} \Big\{ \frac{\mathbb{I}(S_n = s)}{p(S_n | X_n)} \ \prod_{k=1}^K \frac{p(M_{k,n} | s', \overline{A}_{k-1,n},\overline{Y}_{k-1,n}, X_n)}{p(M_{k,n} | s, \overline{A}_{k-1,n},\overline{Y}_{k-1,n}, X_n)} \\
	&- \frac{\mathbb{I}(S_n = s')}{p(S_n | X_n)} \Big\} \ A_{kn} \nonumber
	\end{align*}
}%
Then, in Theorem 2 we analogously define $\tilde{p}(Z)$ as follows:
{\small
	\begin{align*}
	\tilde{p}(Z) \equiv p(X)p^*(S|X;\alpha_s) \prod_{k=1}^K \Big\{ &p^*(M_k|S,\overline{A}_{k-1},\overline{Y}_{k-1},X;\alpha_{m})\\
	&\times p(A_k|H_k) p(Y_k|A_k,H_k) \Big\}.\\
	\end{align*}
}%
In this case we constrain the $S$ and $M_k$ models $\forall k$, the rest of the procedure remaining the same.
Aside from the form of the identifying functional, the proofs of modified versions of Theorem 1 and Theorem 2 are analogous.

{\small
	\bibliographystyle{plainnat}
	\bibliography{references}
}